# Immune System Approaches to Intrusion Detection - A Review

**Jungwon Kim[§], Peter J. Bentley[§], Uwe Aickelin*, Julie Greensmith*,**

**Gianni Tedesco[†], Jamie Twycross***

[§]Department of Computer Science, University College London, UK

j.kim,p.bentley@cs.ucl.ac.uk

*School of Computer Science, University of Nottingham, UK

uxa,jqg,jpt@cs.nott.ac.uk

†Firestorm Development Team, Bradford, UK

gianni@scaramanga.co.uk

## Abstract

The use of artificial immune systems in intrusion detection is an appealing concept for two reasons. Firstly, the human immune system provides the human body with a high level of protection from invading pathogens, in a robust, self-organised and distributed manner. Secondly, current techniques used in computer security are not able to cope with the dynamic and increasingly complex nature of computer systems and their security. It is hoped that biologically inspired approaches in this area, including the use of immune-based systems will be able to meet this challenge. Here we review the algorithms used, the development of the systems and the outcome of their implementation. We provide an introduction and analysis of the key developments within this field, in addition to making suggestions for future research.

# 1 Introduction

One of the central challenges with computer security is determining the difference between normal and potentially harmful activity. For half a century, developers have protected their systems using rules that identify and block specific events. However, the nature of current and future threats in conjunction with ever larger IT systems urgently requires the development of automated and adaptive defensive tools. A promising solution is emerging in the form of biologically inspired computing, and in particular artificial immune systems (AIS). The Human Immune System (HIS) can detect and defend against harmful and previously unseen invaders, so can a similar system be built for our computers? Perhaps, those systems would then have the same beneficial properties as the HIS such as error tolerance, adaptation and self-monitoring [48].

Alongside other techniques for preventing intrusions such as encryption and firewalls, Intrusion Detection Systems (IDS) are another significant method used to help safeguard computer systems. The main goal of these systems is to detect unauthorised use, misuse and abuse of computer systems by both system insiders and external intruders [88]. IDS can be broadly classified into two approaches: anomaly detection and misuse detection.

Considering the above, one can see an analogy between the HIS and IDS. The HIS has both innate and adaptive components to its mechanisms. For example, an innate response is inflammation – the attraction of lyphocytes to the site of an injury and their automatic consumption of dead cells. An adaptive response is a response learned during the lifetime of an organism, such as the production of specific antibodies from carefully maintained populations of B cells. The innate part of the HIS is akin to the misuse detector class of IDS. Similarities can also be drawn between the adaptive immune system and anomaly based IDS. Both the innate HIS and misuse detectors have prior knowledge of attackers and detect them based on this knowledge. Similarly, both the adaptive immune system and anomaly detectors generate new detectors to find previously unknown attackers.

The objective of this review paper (which is an extension of [7]) is to provide an overview of IDS for AIS researchers to identify suitable intrusion detection research problems, and to provide information for IDS researchers about current AIS solutions. Such a review is now important, as a sufficiently large body of research has been amassed to take stock and consider what further avenues should be explored in the future. In the following sections, we briefly introduce the areas of IDS and AIS through the examination of core components and basic definition. The research, development and implementation of immune-inspired IDS are catalogued, and presented in terms of the history and progression of research in the field. The overview of this historical review on this research area shows that it has three major roots, and consequently three distinct philosophies. We provide an extensive survey of various AISs for intrusion detection based on these major roots, in conjunction with indications for future areas of study.

# 2 IDS and AIS Background

This section gives a brief introduction to two distinct fields of study - IDS and AIS, setting the background to and defining the terminology used in the sections that follow. For a detailed discussion readers should consult [96] and [97] for information on IDS, [48] for information regarding immunology and [32] for details specific to AIS. Some of the commonly used terms in this field are given in Appendix 1.

## 2.1 Intrusion Detection Systems

### 2.1.1 Brief Overview of Intrusion Detection Systems

IDS are software systems designed to identify and prevent the misuse of computer networks and systems. There are a number of different ways to classify IDS. Here we focus on two ways: the analysis approach and the placement of the IDS, although there have been recent work [35] [10] on alternative taxonomies. Regarding the former, there are two classes: misuse detection and anomaly detection [96]. The misuse detection approach examines network and system activity for known misuses, usually through some form of pattern-matching algorithm. In contrast, an anomaly detection approach bases its decisions on a profile of normal network or system behaviour, often constructed using statistical or machine learning techniques. Any event that does not conform to this profile is considered anomalous. Many experimental anomaly detection systems exist including [69] [108] [34] and a commercial system also adopts the anomaly detection approach [4].

Each of these approaches offers its own strengths and weaknesses. Misuse-based systems generally have very low false positive rates, which indicate error rates of mistakenly detected non-intrusion cases. For this reason, this approach can be seen at work in the majority of commercial systems, which include [2] [5]. However, they are unable to identify novel or obfuscated attacks, leading to high false negative rates, which represent error rates of missed detection cases. Anomaly-based systems, on the other hand, are able to detect novel attacks but currently produce a large number of false positives. This stems from the inability of current anomaly-based techniques to cope adequately with the fact that in the real world normal, legitimate computer network and system usage changes over time, meaning that any profile of normal behaviour also needs to be dynamic [97].

A second distinction can be made in terms of the placement of the IDS. In this respect IDS are usually divided into host-based and network-based systems [96]. Host-based systems such as tripwire [79] [120] are present on each host that requires monitoring, and collect data concerning the operation of this host, usually log files, network traffic to and from the host, or information on processes running on the host. In contrast, network-based IDSs monitor the network traffic on the network containing the hosts to be protected, and are usually run on a separate machine termed a sensor [99] [91]. Once again, both systems offer the advantages and disadvantages. Host-based systems are able to determine if an attempted attack was indeed successful, and can detect local attacks, privilege escalation attacks and attacks which are encrypted. However, such systems can be difficult to deploy and manage, especially when the number of hosts needing protection is large. Furthermore, these systems are unable to detect attacks against multiple targets of the network. Network-based systems are able to monitor a large number of hosts with relatively low deployment costs, and are able to identify attacks to and from multiple hosts. However, they are unable to detect whether an attempted attack was indeed successful, and are unable to deal with local or encrypted attacks. Hybrid systems, which incorporate host- and network-based elements can offer the best protective capabilities, and systems to protect against attacks from multiple sources are also developed [6] [3].

More advanced systems exist which detect high-level intrusion scenarios through correlation of multiple low-level events. Such systems not only allow for the detection of non-trivial or distributed intrusions spanning multiple events and sources, they can also combine poor quality detection results from misuse and anomaly detectors to produce more reliable results. Approaches have been based on finding statistical similarities between alerts [115], fusing alerts into attack scenarios [28], and knowing the prerequisites and consequences of certain attacks [1]. In [95] it is stated that approaches like these shared a common problem: the IDS can fail to detect an intrusion if the set of reported alerts does not constitute a complete intrusion scenario.

### 2.1.2 IDS Research Problems for AIS

The main objective of this review paper is to introduce suitable intrusion detection problems to AIS researchers. Previously in [88] [81], Kim and Bentley have presented the requirements for an effective network-based IDS. These requirements can be applied not only to a network-based IDS, but to any type of IDS. These requirements are of particular interest because they could be fulfilled by mechanisms inspired by features of the human immune system. Despite research conducted since the original publication of these requirements, no existing IDS model yet satisfies these requirements completely. We summarise these requirements here in order to analyse whether the existing AIS-based IDSs reviewed in this paper have provided some of these functions. The seven requirements reported in [88] are as follows:

- Robustness: it should have multiple detection points with low operational failure rates and which are resilient to attack

- Configurability: it should be able to configure itself easily to the local requirements of each host or each network component. Individual hosts in a network environment are heterogeneous.

- Extendibility: it should be easy to extend the scope of IDS monitoring by and for new hosts easily and simply regardless of operating systems.

- Scalability: it is necessary to achieve reliable scalability to gather and analyse the high-volume of

audit data correctly from distributed hosts.

- Adaptability: it should adjust over time in order to detect dynamically changing network intrusions.

- Global Analysis: in order to detect network intrusions, it should collectively monitor multiple events generated on various hosts to integrate sufficient evidence and to identify the correlation between multiple events.

- Efficiency: it should be simple and lightweight enough to impose a low overhead on the monitored host systems and network.

Readers should note that these are a subset of current IDS requirements. For broader view of IDS research problems and their requirements, readers are advised to refer to [92].

## 2.2 Artificial Immune Systems

### 2.2.1 Brief Overview of Intrusion Detection Systems

The Human Immune System (HIS) protects the body against damage from an extremely large number of harmful bacteria, viruses, parasites and fungi, termed pathogens. It does this largely without prior knowledge of the structure of these pathogens. This property, along with the distributed, self-organised and lightweight nature of the mechanisms by which it achieves this protection [88], has in recent years made it the focus of increased interest within the computer science and intrusion detection communities. Seen from such a perspective, the HIS can be viewed as a form of anomaly detector with very low false positive *and* false negative rates.

An increasing amount of work is being carried out attempting to understand and extract the key mechanisms through which the HIS is able to achieve its detection and protection capabilities. A number of AIS have been built for a wide range of applications including document classification, fraud detection, and network- and host-based intrusion detection [32]. These AIS have met with some success and in many cases have rivalled or bettered existing statistical and machine learning techniques. Two important mechanisms dominate AIS research: network-based models and negative selection models, although this distinction is somewhat artificial as many hybrid models also exist. The first of these mechanisms refers to systems which are largely based on Jerne's idiotypic network theory [70] which recognises that interactions occur between antibodies and antibodies as well as between antibodies and antigens. Negative selection models use the process of non-self matching selection, as seen with T-lymphocytes in the thymus as a method of generating a population of detectors. This latter approach (along with other newer algorithms) has been by far the most popular when building IDS, as can be seen from the work described in the next section.

### 2.2.2 AIS features for IDS

Although not the main objective, we also aim to provide information for IDS researchers about current AIS solutions in this article. In this section, we present AIS features that would be advantageous to a novel IDS. Two previous papers [88] [106] have already covered this topic and here we summarise that work. Kim and Bentley presented three properties of IDSs that satisfy the seven requirements stated above [88] [81]. Another piece of work by Somayaji *et al.* [106] also identifies twelve immune features that are desirable for an effective IDS.

We summarise these AIS features together after eliminating redundant properties:

- Distributed: a distributed IDS supports roubustness, configurability, extendibility and scalability. It is robust since the failure of one local intrusion detection process does not cripple the overall IDS. It is also easy to configure a system since each intrusion detection process can be simply tailored for the local requirements of a specific host. The addition of new intrusion detection processes running on different operating systems does not require modification of existing processes and hence it is extensible. It can also scale better, since the high volume of audit data is distributed amongst many local hosts and is analysed by those hosts.

- Self-Organised: a self-organising IDS provides adaptability and global analysis. Without external management or maintenance, a self-organising IDS automatically detects intrusion signatures which are previously unknown and/or distributed, and eliminates and/or repairs compromised components. Such a system is highly adaptive because there is no need for manual updates of its intrusion signatures as network environments change. Global analysis emerges from the interactions among a large number of varied intrusion detection processes.

- Lightweight: a lightweight IDS supports efficiency and dynamic features. A lightweight IDS does not impose a large overhead on a system or place a heavy burden on CPU and I/O. It places minimal work on each component of the IDS. The primary functions of hosts and networks are not adversely affected by the monitoring. It also dynamically covers intrusion and non-intrusion pattern spaces at any given time rather than maintaining entire intrusion and non-intrusion patterns.

- Multi-Layered: a multi-layered IDS increases robustness. The failure of one layer defence does not necessarily allow an entire system to be compromised. While a distributed IDS allocates intrusion detection processes across several hosts, a multi-layered IDS places different levels of sensors at one monitoring place.

- Diverse: a diverse IDS provides robustness. A variety of different intrusion detection processes spread across hosts will slow an attack that has successfully compromised one or more hosts. This is because an understanding of the intrusion process at one site provides limited or no information on intrusion processes at other sites.

- Disposable: a disposable IDS increases robustness, extendibility and configurablity. A disposable IDS does not depend on any single component. Any component can be easily and automatically replaced with other components.

These properties are important in an effective IDS, as well as being established properties of the HIS. The individual immune mechanisms that allow each AIS to have these properties are presented in the next section 3. Later in section 4, we summarise whether AISs introduced in section 3 have indeed these properties and which immune algorithms employed by AISs contribute to obtaining these properties. This leads to an analysis of how such features support the IDS requirements presented in section 2.1. The remainder of this paper reviews various systems that provide some of these properties, by adopting artificial components inspired by their biological counterparts in the immune system.

# 3 Immune System Approaches to IDS

In this section, we begin the in-depth review of work relating to the application of AIS to the problem of intrusion detection. The work reviewed in this section is organized by the history and progression of research in the field, see the phylogenetic tree of papers given in figure 1.

The figure shows that work in this area has three major roots, and consequently three distinct philosophies:

1. methods inspired by the immune system that employ conventional algorithms, for example, IBM's virus detector [76]
2. the negative selection paradigm as introduced by Forrest [106][45]
3. approaches that exploit the Danger Theory [93]

In addition, there are several younger methods such as AINET [27] and immunocomputing [94] that continue to grow in popularity. The following subsections explore all the philosophies behind AIS for intrusion detection and analyse their successes and capabilities to date.

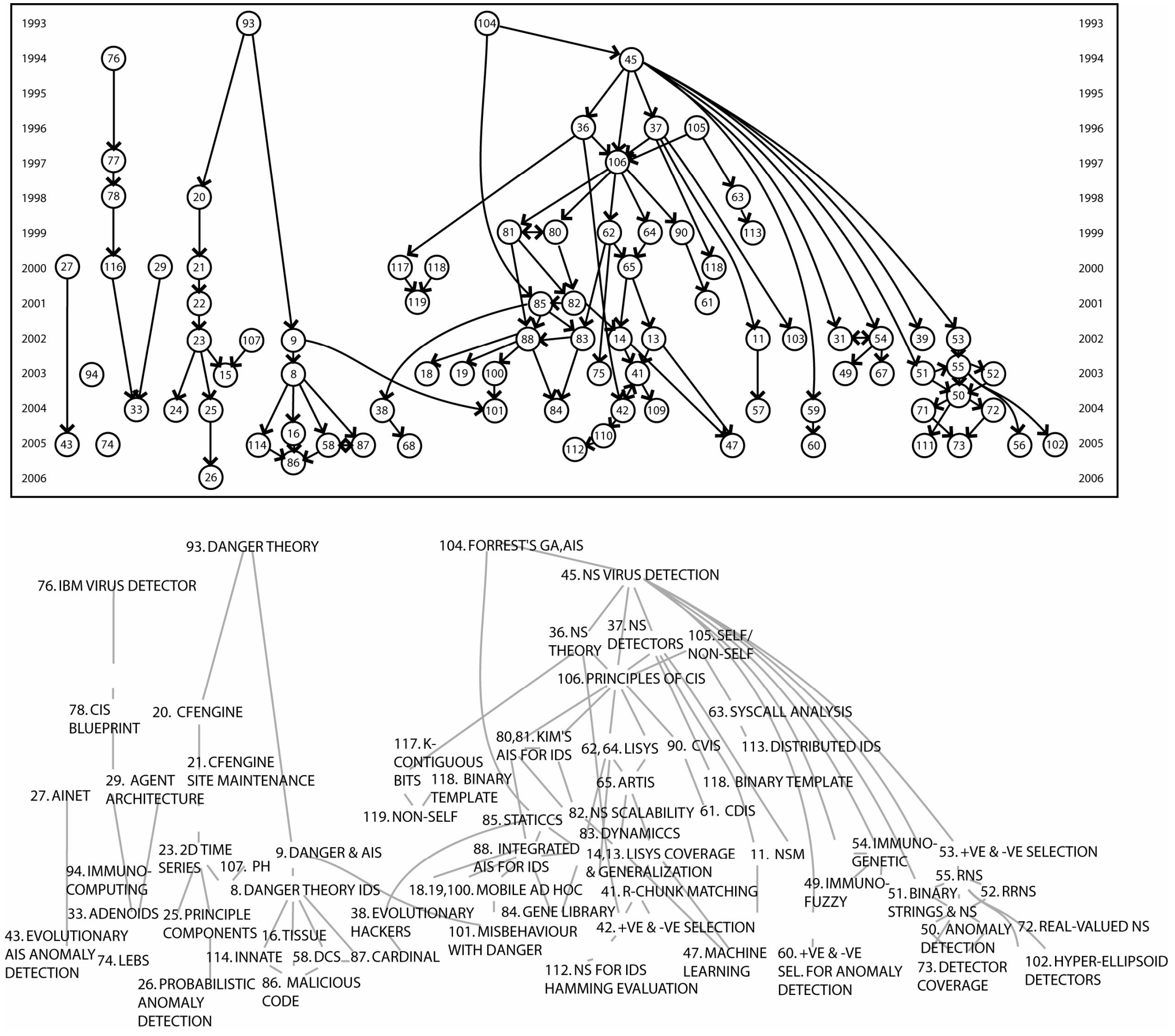

**Figure 1 Phylogenetic tree of AIS approaches to IDS. Detailed organization arranged by date and content (top). Key summarizing major papers (bottom).**

## 3.1 Exploitation of Conventional Algorithms in AIS

As will be described in subsequent sections, most AIS research focuses on the development of specialised AIS algorithms inspired by specific theories of the HIS such as negative selection or the danger theory. But before these fields of research had begun, a virus detection system had been developed by Kephart *et al.* at the IBM research centre [76] [77] [78] following an alternative approach (see figure 1). Kephart *et al.* identified some traits of the HIS that make it attractive for virus detection purposes and implemented them using established algorithms. They designed their AIS with five major stages, each inspired by the HIS. Taking one of these five stages as an example, authors claimed that the first stage of their AIS, which detected a previously unknown virus on a user's computer, is a useful trait of the innate human immune system. Since the innate HIS provides a non-specific immune response, Kephart *et al.* viewed the detection of previously unknown viruses in a generic way as an equivalent task in their artificial system. The actual implementation of this innate immune response was carried out by two established algorithms: generic disinfection techniques and neural networks, which were used to build a generic classifier. Similarly, the other four stages of their AIS were also understood to be functionally comparable with some sub-procedures of the HIS, but their implementation was completed using other more conventional algorithms.

The AIS described by Kephart [76] [77] [78] was one of the earliest attempts of applying HIS mechanisms to intrusion detection. It focused on the automatic detection of computer viruses and worms. As interconnectivity of computer systems increases, computer viruses are able to spread more quickly and traditional signature-based approaches, which involve the manual creation and distribution of signatures, become less effective. Hence the authors were interested in creating a system which was able to automatically detect and respond to viruses. Their proposed system first detected viruses using either fuzzy matching from a pre-existing signature of viruses, or through the use of integrity monitors which monitored key system binaries and data files for changes. In order to decrease the potential for false positives in the system, if a suspected virus was detected it was enticed by the system to infect a set of decoy programs whose sole function was to become infected. If such a decoy was infected then it was almost certain that the detected program was a virus. In this case, a proprietary algorithm, not described in the paper, was used to automatically extract a signature for the program, and infected binaries were cleaned, once again using a proprietary algorithm not described in the paper. In order to reduce the rapid spread of viruses across networks, systems found to be infected contacted neighbouring systems and transfered their signature databases to these systems. No details of testing and performance were given by the authors, who claimed that some of the mechanisms were already employed in a commercial product [116]. It seems likely that the only feature of immune systems exploited significantly was self-organisation.

Dasgupta *et al.* [29] also proposed an alternative immunity-based IDS framework that applied a multi-agent architecture (see figure 1). This immunity-based IDS framework followed the multi-level detection feature of the HIS. The multi-agents residing within this model monitored systems at various levels in a hierarchical manner, activated warning signals, communicated their local warning signals and made decisions based on collected local warning signals. In order for the proposed IDS to reduce false warnings, this model emphasised the importance of collective decisions. The ART-2 NN was employed to detect anomalies of all monitoring levels and fuzzy logic was proposed to combine four different levels of warnings into a final threat warning [30]. Hence, like Kephart's system, the AIS framework proposed in [29] used existing algorithms organized in a manner inspired by the multi-level detection features of the HIS.

The recent work by de Paula *et al.* [33] proposed another AIS based IDS called ADENOIDS. This novel AIS conforms to the philosophy taken by Dasgupta *et al.* [29] and Kephart *et al.* [76] [77] [78]. de Paula *et al.* introduced eight different components taken from the innate and the adaptive immune system. From the innate immune system, the *evidence-based detector* is responsible for detecting intrusions based on clear evidence such as a security policy violation. The *innate response agent* reacts to attacks detected by the evidence-based detector. Their responses, such as limiting bandwidth or disk access, are limited and general like the reactions of the innate immune system. The *behaviour-based detector*, which is an anomaly detector, is initiated only when it receives co-stimulation signals, which are the detection restuls of the evidence-based detector. Like the adaptive immune system, the *signature extractor* extracts signatures of detected attacks and has a learning mechanism which allows attack signatures to mature. Some of the matured attack signatures are kept at the *knowledge-based detector* which corresponds to the adaptive immune memory. The signature extractor activates the *response generator* and the *adaptive response agent*. The response generator decides the types of responses and the adaptive response agent performs the selected responses. Altogether, de Paula *et al.* attempted to identify and understand useful processes of the HIS, and to see how these can help with devising new IDS architectures. However, they did not attempt to implement the processes using the mechanism of the HIS, only to mimic it at a high level of abstraction.

While the use of existing algorithms modified in some way to resemble an immune mechanism may improve reliability in some cases, it could be argued that they do not allow all of the possible features of immune systems to be exploited. Some methods were certainly multi-layered and diverse, but the systems often did not have disposable elements or significant properties of self-organisation or being distributed.

## 3.2 Negative Selection Approaches

### 3.2.1 Negative Selection algorithm overview

While some success has been made using conventional algorithms in a manner inspired by the human immune system, IDS researchers were quickly attracted to one specific aspect of the immune system: negative selection in the T-cell maturation process [45]. Negative selection eliminates inappropriate and immature T-cells that bind to self antigens. This allows the HIS to detect non-self antigens without mistakenly detecting self-antigens.

Forrest *et al.* [104][45] [106] were the first to propose a novel negative selection algorithm mimicking this process (see figure 1). They considered the negative selection process of the HIS to be a sophisticated anomaly detection method. This process does not define specific harmful cells to be detected and thus allows the HIS to be able to detect previously unseen harmful cells. The algorithm consists of three phases: defining self, generating detectors and monitoring the occurrence of anomalies. In the first phase, it defines `self' in the same way that other anomaly detection approaches establish the normal behaviour patterns of a monitored system. It regards the profiled normal patterns as `self' patterns. In the second phase, it generates a number of random patterns that are compared to each self-pattern defined in the first phase. If any randomly generated pattern matches a self-pattern, this pattern fails to become a detector and thus it is removed. Otherwise, it becomes a detector pattern and monitors subsequent profiled patterns of the monitored system. During the monitoring stage, if a detector pattern matches any newly profiled pattern, it is then considered that new anomaly must have occurred in the monitored system.

In [36], D'haeseleer *et al.* highlight a number of the NS algorithm features that distinguish it from other intrusion detection approaches. They are as follows:

- No prior knowledge of intrusions is required: this permits the NS algorithm to detect previously unknown intrusions.
- Detection is probabilistic, but tunable: the NS algorithm allows a user to tune an expected detection rate by setting the number of generated detectors, which is appropriate in terms of generation, storage and monitoring costs.
- Detection is inherently distributable: each detector can detect an anomaly independently without communication between detectors.
- Detection is local: each detector can detect any change on small sections of data. This contrasts with the other classical change detection approaches, such as checksum methods, which need an entire data set for detection. In addition, the detection of an individual detector can pinpoint where a change arises.
- The detector set at each site can be unique: this increases the robustness of IDS. When one host is compromised, this does not offer an intruder an easier opportunity to compromise the other hosts. This is because the disclosure of detectors at one site provides no information of detectors at different sites.
- The self set and the detector set are mutually protective: detectors can monitor self data as well as themselves for change.

The negative selection (NS) based AIS for detecting computer viruses was the first successful piece of work using the immunity concept for detecting harmful autonomous agents in the computing environment [105]. Since this first success, the Adaptive Computation Group at the University of New

Mexico, headed by Stephanie Forrest, has been instrumental in the development of IDS, employing concepts and algorithms from the field of AIS. Not limiting the potential strengths of AIS on a simple virus detection problem, the following work [106] drew on many principles of the HIS that would guide the design of future AIS for IDS (see figure 1).

In [106], Forrest *et al.*'s work identified the features of the HIS desirable for imperfect, uncontrolled and open environments, which reflected the current computer security environment. The HIS features identified by this work were *distributability, multi-layeredness, diversity, disposability, autonomy, adaptability, dynamic coverage, anomaly detection, identity via behaviour, no trusted components* and *imperfect detection* [106]. The authors underlined the key advantageous trait of the HIS as its unique capability to incorporate these properties all together. They pointed out that there was no single computer security system that was equipped with more then a few of these properties despite some of the properties being already adopted by current computer systems in isolation. It was then shown that four possible AIS architectures exist where the desirable properties could be implemented within. They were protecting static data (A1), protecting active process on a single host (A2), protecting a network of mutually trusting computers (A3), and protecting a network of mutually trusting disposable computers (A4).

The attractive features of negative selection have led a number of researches to employ the NS algorithm for the study of IDS (the major branches for NS shown in figure 1 illustrates how widespread such research has become). In the following sections, we present promising results of the NS algorithm shown in applications within security. We will also focus on how researchers have tried to resolve the two notable but intrinsic problems of the NS algorithm - scalability and coverage.

### 3.2.2 Negative Selection anomaly detection for security

Early work by the Adaptive Computation Group at the University of New Mexico [45] viewed virus detection as a self-nonself discrimination problem within a computer. They regarded monitoring targets (such as legal user activities, legal application usage activities, uncorrupted data, etc.) as self and expected the NS algorithm to discriminate them from others (such as illegal user activities, illegal application usage activities, virus infected data, etc.). In this work, Forrest *et al.* randomly generated binary string detectors and selected the subset which did not match to self strings from a standard binary executable .com file. The experimental results showed that the NS algorithm obtained a 100% detection rate under a relatively small scale problem: with 125 detectors when an infected file was encoded by 655 binary strings each string having 32 bits.

More recent work by Hofmeyr and Forrest [62] [64] involved the development of an AIS for network intrusion detection, called LYSIS. LYSIS implements the AIS architecture called `ARTIS' described in [65]. It employs the NS algorithm for binary detector generation and various features of the HIS such as *activation threshold, life span, memory detectors, costimulation, tolerisation period* and a *decay rate*, in order for it to monitor *dynamic* self and non-self behaviours. LYSIS is network-based and examines TCP connections, classifying normal connections as self, and everything else as non-self. In order to perform this, LYSIS extracts a datapath triple, which is a source host IP address, a destination host IP address and a TCP service (port) number from TCIP/IP packet headers. This datapath is used as input data to build self-profiles. Detectors in the form of binary strings which do not match to self-profiles for a tolerisation period are generated using NS. They are subsequently matched against sniffed triplets from the network using an r-contiguous bit matching scheme[1]. If a detector matches a number of strings above an activation threshold, an alarm is raised. Detectors that produce many alarms are promoted to memory cells with a lower activation threshold to form a secondary response system. Generated detectors monitor a network for their life span periods. Co-stimulation is provided by a user confirming if an alert is genuine, which reinforces true positives. The activation threshold is set according to an adaptive mechanism involving many local activation thresholds, based on match counts of detectors.

LYSIS was tested by using it to monitor 50 hosts in a local area network (LAN), where each host in the LAN independently generated detectors and monitored new traffic. LYSIS was tested on seven intrusions and showed promising results. However, the limited input data suggests that future research may be necessary to evaluate whether LYSIS is able to detect more diverse intrusions then those used in [62] [64].

---

[1] R-contiguous matching measures the similarity between two binary strings by counting contiguously matching bits. Together with Hamming distance, r-continuous matching is initially suggested for detector matching methods of the NS algorithm [45][62].

Balthrop *et al.* [14] [47] provided an in-depth analysis of the LISYS immune-based IDS [62]. In this analysis, the adaptive mechanisms of the LISYS immune-based IDS were examined with respect to machine-learning (ML) counterparts, and the contribution of an each individual component was quantified. For this analysis, the authors collected new data from an internal restricted network of computers that the authors themselves controlled. A total of six internal hosts were connected to the Internet through a single Linux box, where firewall, router and masquerading server ran. After the week-long normal period ended, several attacks were performed through the use of Nessus [98] for two days. This study aimed to show how LISYS accomplished its success as an effective IDS and how individual mechanisms of the LISYS could be combined to other ML mechanisms. The results of this work showed that many individual components of the LISYS could valuable in other challenging ML problems such as one-class learning, concept drift and on-line learning. Specifically, co-stimulation together with memory detectors provided the LISYS with the extended on-line and one-class learning mechanisms, and rolling-coverage handled the concept-drift. In addition, activation thresholds and sensitivity levels contributed to reduce false positives, and the incorporation of r-chunks and permutation masking also reduced false positives and increased true positives [13]. Together these mechanisms provided an impressive array of immune features (not beaten by any other approach to date): distributed, self-organised, lightweight, diverse and disposable.

Another approach, called Computer Virus Immune System (CVIS) [90] [61] by the US AFIT team, was developed as a part of Computer Defence Immune System (CDIS). CVIS has features including detected virus analysis, repair of infected files and dissemination of analysis results to other local systems. In addition, CVIS was designed to operate under a distributed environment using autonomous agents but the reported test results of CVIS [90] [61] were limited to evaluation focused on a local host based implementation. The viruses tested were the TIMID virus, which infects .com files only within a local directory. The test reports showed the sensitivity of detection and error results depending on different matching thresholds. By setting appropriate parameters, this system was able to show a detection rate of up to 89%. One serious problem found from these tests was scalability. Notably CVIS required approximately 1.05 years for generated antibodies to scan an 8GB hard disk drive.

Harmer *et al.* extended CVIS further to detect network intrusions [90] [61]. The extended system, named CDIS, also used the NS algorithm with some novel ideas that had been introduced in LYSIS such as life span, activation threshold and co-stimulation. In contrast to LYSIS, CDIS chooses 28 features of TCP packet headers and 16 features of UDP and ICMP packet headers as its input self data. For detector string generation, CDIS randomly selects the network protocol and chooses between two and seven features of that selected protocol. For the selected protocol features, CDIS generates the values of these features randomly. As a new way of reducing the computing time to generate antibody strings, CDIS adopts an affinity maturation process. This process aims to optimise each detector to cover the non-self space as much as possible without matching any self string. In order to perform this, CDIS uses a genetic algorithm and its fitness function is defined as the growth rate of non-self space coverage by each antibody. The test results showed that CDIS was able to detect simulated intrusions without serious self detection errors. The results also verified that the co-stimulation and affinity maturation help CDIS to reduce both FP and FN error rates. However, it was found that the affinity maturation required far too much computation time to be applied to the second, larger, data set. In addition, the authors also pointed out that the high detection rates with low error rates might have been obtained because the simulated intrusions were limited. As with a number of IDS more extensive tests with more varied intrusions are required in order to fully validate these techniques.

In more recent work, Le Boudec and Sarafijanovic [18], [19] [100] built an immune-based system to detect misbehaving nodes in a mobile ad-hoc network. (These are wireless networks in which each end-user system, termed a node, acts as both a client and router. As nodes act as routers, their proper functioning is essential for the transmission of information across the network.) The authors considered a node to be functioning correctly if it adhered to the rules laid down by the common protocol used to route information, in their case the Dynamic Source Routing (DSR) protocol. Each node in the network monitored its neighbouring nodes and collected one DSR protocol trace per monitored neighbour. Four sequences of DSR protocol events were sampled over fixed, discrete time intervals to create a series of data sets. This created a binary antigenic representation in which each of the four genes recorded the frequency of their four sequences of protocol events. The NS algorithm was then used with the generated antigens and a set of uniformly randomly-generated antibodies to eliminate any antibodies which matched, using an exact matching function. Once a mature set of detectors had been generated, these antibodies were used to monitor further traffic from the node and, if they matched antigens from the node, classify it as suspicious. While more work needs to be performed before this approach can be implemented on a real ad-hoc network, it has the potential to be distributed, self-organised, disposable and must be light-weight or the nodes would not be able to operate effectively. This is also one of the first attempts to explore the use of immune algorithms in the rapidly-growing area of ubiquitous computing – an area that may be highly appropriate for immune algorithms, given the dynamic, distributed nature of such networks.

**Comment [ucacjki1]:** It seems to be a sudden to talk about a ubiquitous computing here. This part might be changed as… "This is also one of the first attempts to explore the use of immune algorithms in a mobile ad-hoc network (MANET) environment. The MANET environment has been recently becoming a significant research area with the rapidly-growing new application such as ubiquitous computing. The MANET together with sensor networks, P2P networks, and mesh networks become the major infra-structure of ubiquitous computing and these networks may be highly appropriate for immune algorithms to operate, given the dynamic, distributed nature of such networks".

### 3.2.3 Scalability of the NS algorithm

In their 1996 work, D'haeseleer *et al.* [37] developed binary detector generation algorithms using a linear-time algorithm that guaranteed a linear time of detector generation, and a greedy algorithm that generated non-redundant detectors in linear time. Later, Singh [103] extended the greedy algorithm that was able to handle strings with a large alphabet cardinality, and Wierzchon [118] introduced the binary template that helps the NS algorithm to generate non-redundant detectors more efficiently. However, these algorithms require the use of a contiguous bit matching method and they trade space complexity for time complexity. For these reasons, Hofmeyr's LYSIS (see pp32 in [62]), the network-based IDS operating on a real environment, and Kim and Bentley's work [82] [88] used the original negative selection algorithm [45] for binary detector generation.

In order to investigate the feasibility of the NS algorithm in a real network environment, Kim and Bentley [82] [88] studied the problem of scalability of the NS algorithm. For this study, they used TCP packet headers covering around 20 minutes and containing five specified attacks. A total of 33 different attributes were extracted describing a specific network connection. These attributes contained the following information: connection identifier, known port vulnerabilities, 3-way handshake details and traffic intensity. To encode this data into self strings, the alphabet cardinality used was 10, which is significantly larger than the binary encoding that is usually employed by other NS algorithms [45][62] [61]. For detector matching, the r-contiguous matching method was used. Non-self detection rates for the various attacks were recorded as less than 16% so the detector coverage in this case was not sufficient. It was estimated that for an 80% detection rate it would take 1,429 years to produce a detector set large enough to achieve this kind of accuracy, using just 20 minutes worth of data, and $6*10^8$ detectors would be needed. From these results, the authors concluded that the NS algorithm produced poor performance due to scaling issues on real-world problems.

The following work by Kim and Bentley [88] [83] introduced a dynamic clonal selection algorithm (DynamiCS) that controlled the proliferation and extinction of detectors within IDS. DynamiCS was an integrated model that combined negative selection, clonal selection and gene library evolution. It built on a simplified version of LISYS [62] and hence employed tolerisation periods, costimulation, affinity maturation, life span and memory detectors. The authors showed that DynamiCS was able to incrementally learn globally converged normal behaviours by being exposed to only a small subset of self antigens at one time. In [88] [84], the extended DynamiCS eliminated memory detectors when they showed poor self-tolerance to new antigens. The experimental results showed that deletion of memory detectors based on their self-antigen detection dramatically decreased high false positive rates. In order to reduce the large amount of costimulation (confirmation to be performed by a human security officer), DynamiCS employed a "virtual gene library", made from mutations of deleted memory detectors [88] [84]. The new extension was tested to determine whether it gained high true positive detection rates without increasing the amount of costimulation as the result of gene library evolution. With gene library evolution, DynamiCS showed a smaller true positive rate drop when antigens were presented from the same antigen cluster. This was because the mutants of previously deleted memory detectors, having survived the negative selection stage, were likely to have some non-self antigen information without patterns matching self antigens. DynamiCS exploited several immune features such as being self-organised and diverse. However, the aforementioned research by Kim and Bentley [85] [83] [84] was tested only on a small size of data sets selected from the UCI repository machine learning database [17].

Following the criticism by Kim and Bentley in [82] regarding scaling and false positives, Balthrop *et al.* [14] provided some explanation of the results reported in [82]. In particular, they criticised the choice of the matching threshold for r-contiguous matching function and the cardinality of alphabet used for the detector genotype. Balthrop *et al.* suggested that jumping from a value of 4 to 9 for the matching threshold value is not an ideal approach to tune the performance sensitive parameter. However, Kim and Bentley [82] clarified that the matching threshold value of 4 was tried and then, in a series of experiments, the value was gradually increased until it finally generated a single valid detector in a reasonable time, (approximately 70 seconds CPU time). They also explained that the choice of large alphabet cardinality for the experiments performed in [82] was necessary because of the much larger number of fields and their possible values to be presented then ones used in [62] and [14]. Nevertheless, Balthrop *et al.* make a strong point in [14] that the problem could lie in the representation and the r-continuous match rule, not in the negative selection process itself.

Balthrop *et al.* [14] also introduced an improvement to r-contiguous matching called an r-chunk scheme. In this scheme, only r contiguous bits of the whole detector are specified (known as the *window*), with the remaining becoming wild-cards and thus the partial matching is performed. Subsequent work by Esponda *et al.* [42] reported that the r-chunk matching requires $O(t\,|S|)$ time and $O(t2^r)$ space where t is the number of windows in a given string, S is a collected self set, and r is the length of the matching chunk. In other words, r-chunk matching shows linear-time complexity against the number of self-patterns and windows but requires more space compared to the original NS algorithm. Recent work by Stibor *et al.* [109] employed a hashtable data structure for generating r-chunk matching detectors. The elements in the hashtable are a composite of a r-bit chunk string and a position of the r-chunk within a given detector string. The hashtable key returns the element as a boolean value indicating whether the corresponding r-chunk matches any self string or not. Thus, valid detectors are simply returned elements of the hashtable when the key value is *true*. However, this modification still shows the time complexity exponential to r which is $O(|\Sigma|^r)$ when $\Sigma$ is the alphabet cardinality of a detector string.

More recently, Stibor [110] [112] et al has shown that the generated detector set underfits exponentially for small value *r*. Underfitting behaviour leads a user to set the matching threshold value r near *l*. However, this verifies that the detector generation using the negative selection with r-chunk matching infeasible since all the proposed variants of the negative selection algorithm have a runtime complexity which is exponential in *r*.

While the above work attempted to reduce detector generation time using different matching methods, Ayara *et al.* [11] proposed a modification of the original NS algorithm, which used somatic hypermutation[2] (The original NS algorithm was termed the exhaustive NS algorithm in their paper). This new algorithm was called negative selection mutation (NSM) and performed a guided mutation on the detector which matched self data during the detector generation process. The specific parts of a detector used to match the bits to a self-string were targeted for mutation. The mutation rate was dynamically set according to the affinity between a detector and a self string: the greater the affinity, the higher the mutation rate. The number of mutations performed on the same candidate detector were restricted. The authors compared the NSM with the exhaustive NS through the tests performed on randomly generated 8-bit self data. The results illustrated that the two algorithms showed similar time complexity and detection rates with no statistical significant differences. However, the authors argued that these results were likely to be caused by the nature of randomly generated self data. This was because the executed mutations resulted in the detectors being pushed towards or away from a self-string with an equal probability. They also added that in unpublished following work which tested the NSM on structured self-data they showed better performance in terms of time complexity.

In recent work by Gonzalez and Cannady [57], the authors improved the NSM algorithm by adopting the self-adaptive strategy of evolutionary algorithms to control the mutation rate [40]. This strategy determines a mutation rate at every generation by selecting the standard deviation from the fittest detectors selected via a tournament selection, multiplied by Gaussian noise. A comparison with the NSM algorithm showed that the new algorithm performed better with respect to higher detection rates, lower false detection rates, and computation time taken. Other work by Hang and Dai [59] used common schemata of self data in order to reduce the non-self search space. The authors let the co-evolutionary algorithm run to find common schema that exist among self-patterns. Then, randomly generated binary detectors were matched against the common schema of self data only, instead of entire self strings. The motivation behind this work was that it could reduce the number of self strings, which would save computation time and still find a reasonable quality of detectors. However, the algorithm was only tested on the well-known *Iris* data set and no investigation on the time complexity compared to the exhaustive NS algorithm was reported. Moreover, the authors did not discuss newly arising problems such as the escalation of false positive error rates possibly caused by simplification of the self space.

---

[2] Somatic hypermutation is the occurrence of a high level of mutation in the variable regions of B-lymphocytes with the possible purpose of increasing the binding affinity to antigens.

After observing the scaling problem of the NS algorithm in [82] [88], Kim and Bentley [85] suggested the need to re-define the role of the NS stage within a network-based IDS, and to design a more applicable NS algorithm according to that new role. The new algorithm, called the *static clonal selection* algorithm (StatiCS) used the clonal selection process of the HIS with a NS operator. It let detectors evolve towards the non-self patterns hidden in the collected non-self data and the NS operator acted as a filter for invalid detectors, not the generation of competent detectors. This algorithm was developed especially for the purpose of building a misuse detector in a more efficient way. StatiCS was an enhancement to the clonal selection algorithm introduced in [104]. In order for StatiCS to be applied to the network intrusion detection problem, a modified representation of detectors and a matching function were introduced. A binary string was used as detector genotype and a conjunctive rule was employed as detector phenotype. The new phenotype representation removed the matching threshold parameter by allowing an "OR" operator in a genotype-phenotype mapping. The modified phenotype representation no longer required the arbitrary choice of a parameter value that significantly affected the detection rate. In addition, this phenotype allowed a detector to match more then one specific antigen and thus it still had the lightweight feature, originating from approximate binding. Furthermore, the detector phenotypes used in this work had a larger degree of intelligibility. Nevertheless, StatiCS remains unproven for large-scale network intrusion detection.

### 3.2.4 Coverage of the NS algorithm

With the initial success of the NS algorithm in virus detection, the researchers at the University of New Mexico studied the theoretical aspects of the NS algorithm [37] [36]. In this work, the authors explained the concept of holes existing in the NS algorithm. Depending on matching methods and strings used in the NS algorithm, there exist non-self strings called *holes* that are not covered by a complete detector repertoire. Figure 2 illustrates the existence of holes in a self and a non-self space comprised by self and detector strings.

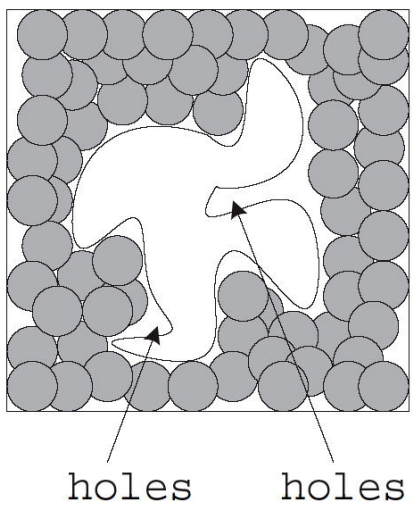

**Figure 2: The existence of holes. Each dark circle represents a detector and a gray shape in the middle is self-antigen data. The size of the dark circles reflects the generality of detectors. Since all the detectors have an identical radii, and the detectors are too general to match some non-self subspaces without matching self antigen data, there inevitably exist holes [62].**

As illustrated in Figure 2, this problem arises from the adoption of symmetrical string matching and its generality. The existence of holes determines a lower bound on a false negative error rate. In order to remedy this problem, Hofmeyr [62] [64] introduced the *permutation mask*, inspired by the function of the important MHC (Major Histocompatibility Complex) class of molecules.

In the HIS, there are a number of classes of MHC, most noteworthy being class I and II. The presentation of peptides in combination is vital for the presentation of antigens to lymphocytes. It has been shown that MHC molecules are highly polymorphic and in many cases are unique for individuals. Therefore, the diversity of MHC classes provides a better chance for the successful detection of different types of antigens. Inspired by this insight, the permutation mask lets the NS algorithm randomly permutate the binary bits of generated detectors. As a consequence, it has an additional set of detectors with different representations reflecting an identical non-self space. Different representations would have different holes in a non-self space and hence the union of non-self space coverages by multiple sets of detectors are likely to result in reducing the number of holes.

By introducing the permutation mask, Hofmeyr [62] [64] demonstrated improved detection results by up to a factor of 3, especially when LISYS attempted to detect a non-self string close to a self-string in a search space. Kaers *et al.* [75] applied the permutation mask on their fuzzy rule detectors and tested it on three data sets from the UCI repository machine learning databases [17]. Their test results also produced improved detection rates. Balthrop *et al.* [13] also investigated the effect of the permutation mask used by the simplified version of LISYS, which employs r-chunk matching. From this investigation, they found that the incorporation of r-chunks and permutation masking reduced false positives and increased true positives. Additionally, they found that varying r had little effect, unlike with full-length detectors. As the r-chunks scheme performed remarkably well the authors investigated it further, and subsequently found that the dramatic increase in performance was in part due to the configuration of their test network. Nevertheless, it still outperformed the full-length detector scheme.

In later work, Esponda *et al.* [42] formally showed that r-contiguous matching augmented by the permutation mask is able to cover a larger space recognised by Hamming distance matching. Their study also showed that there are still non-self strings not detected by r-contiguous matching augmented by the permutation mask. (Wierzchon [118] [117] [119] had also presented the number of existing holes and the lower bound of a detection failure error rate based on his new algorithm, another modification the NS algorithm using a binary template. This analysis, however, was limited to using r-contiguous matching on static data.) More importantly, Esponda *et al.'s* work [42] provided a formal framework for positive and negative detection schemes that analysed the trade-offs between two schemes. Firstly, they emphasised the necessity of generalising a set of strings to be recognised by the adopted matching methods, in order to discriminate non-self patterns from self patterns in a realistic amount of time. This understanding calls for a partial matching method to generalise detector strings. In order to explain the relation between a partial matching and string generalisation, the property termed *crossover closure* is introduced. When all the possible sliding windows of each string existing in a universal string set exactly match the corresponding windows of some self-strings, it is said that the self-string set is *closed under crossover closure*. The authors used this property to characterise two matching methods: r-contiguous and r-chunk matching. It was formally proven that when these two matching methods were employed by two different detecting approaches - negative disjunctive and positive conjunctive, both cannot recognise all string sets under crossover closure, and both can recognise some sets not under crossover closure. It was additionally shown that the string space recognised by negative disjunctive detection is equivalent to the one covered by positive conjunction detection. The authors continued to estimate the expected number of detectors when r-chunk is applied for negative disjunctive or positive conjunctive detection. For the first time, an estimation was developed for how many self-strings are required for either negative disjunctive detection or positive conjunctive detection to be computationally advantageous. Moreover, the authors approximated the number of holes as a function of self-strings coupled with a string length and a size of r. An interesting observation from this approximation is that the number of holes decreases as more self strings are added, when the size of r is relatively close to the string length. Further work by the same authors [41] extended the similar formal analysis to the case when non-overlapping sliding windows are used for r-chunk matching.

Whist the above work focused on the improvement of binary string matching and coverage, Gonzalez [50] extensively studied various detector representation schemes of the NS algorithm (see figure 1). The original NS algorithm employed a binary representation of given data and detectors. Gonzalez *et al.* [50] [51] showed the limitations of the NS algorithm which are possibly caused by the binary representation scheme. They argued that matching rules between two binary strings cannot represent a good generalisation of a self space and thus a generated detector set shows poor coverage of a non-self space. The main reason for this problem, the authors explained, was that the affinity relation between two binary strings represented by binary matching rules cannot capture the affinity relation between two data examples in a given problem space. To verify this argument, the authors visualised the coverage space of a single detector on a problem space illustrated by two-dimensional real values, and measured detection and false alarm rates by applying different binary matching rules and matching thresholds. The results of these assessments supported their claim.

Subsequent research by Dasgupta and Gonzalez [31] [54] [50] examined whether the fundamental idea of NS is advantageous for the network intrusion detection problem. They compared a negative characterisation approach to positive characterisation. The positive approach focused on generating rules covering a self-space and detected anomalies by monitoring events that matched no self rules. Their implementation of the positive selection algorithm used a k-dimensional tree, giving a quick nearest neighbour search. On the other hand, their negative characterisation approach employed a genetic algorithm in order to generate detector rules covering niches of a non-self space. While the NS algorithm generated detectors covering the same radii of clusters in non-self space, Dasgupta and Gonzalez used the GA to let detectors evolve to cover generalised niches that had various radii. In order for detector rules to evolve, the fitness function was defined by the volume of non-self space covered by detector rules after a penalty having been applied according to the number of matching self examples. To test the system they used a small subset of the 1999 Lincoln Labs outside tcpdump datasets [89]. Overall, the best detection rates they found were 95% and 85% for positive and NS respectively. They concluded that it is possible to use NS for IDS, and that in their time series analysis, the choice of time window was imperative.

Different work using the same algorithm to detect intrusions on mobile ad hoc networks was reported in [67]. Dasgupta and Gonzalez extended the negative characterisation approach to generate detectors represented by fuzzy rules [50] [49]. Further experiments on the same data set showed the advantages of using fuzzy rules to represent detectors. Specifically, they provided better definition of the boundary between a self and a non-self space, they showed an improved detection accuracy because of the reduction of a search space due to the fuzzy representation and they also generated a more compact representation of a non-self space by reducing the number of detectors.

In [55] [50] [53], Gonzalez and Dasgupta developed the real-valued negative selection (RNS) algorithm. The RNS algorithm employs two distinctive features: the use of real-value representation and hybridising the NS algorithm with a classifier. The limitation of binary representations used by the NS algorithm motivated them to propose a new detector generation algorithm coupled with a novel matching function. In addition to the NS approach, classifier-based supervised learning paradigms have also been suggested as a possible avenue of exploration by IDS researchers [96]. However, a practical difficulty in using a classifier for IDS, namely gathering an extensive set of non-self data, has attracted many researchers to employ the NS algorithm as an anomaly detector. In order to resolve these problems faced by a conventional classifier or the NS algorithm, Gonzalez and Dasgupta used a combination of the two: the NS algorithm to generate artificial non-self data examples and a classifier to learn the non-self space from these examples.

The RNS algorithm uses n-dimensional vectors as detectors. Detectors have a radius r, in other words they represent hyper-spheres in combinations with a fuzzy Euclidean matching function. In training, detectors are generated randomly and then moved to both maximise the coverage of a non-self space and to minimise the coverage of a self space. Detectors match if the median distance to their k-nearest neighbours is less then r and matching detectors are discarded. Surviving detectors are then provided to a multi-layer perceptron classifier trained with back-propagation. From this work, the authors concluded that scaling is not a problem in NS when real values are used rather than binary and r-continuous matching. They also concluded that NS could train their classifier effectively without providing non-self data collected from a real environment. The latest work by Gonzalez et al [56] hybridised the RNS algorithm with a Self-Organizing Map (SOM). This work attempted to visualize anomalies in 2-dimensional map. In contrast, the latest work by Ji and Dasgupta [71] [72] further extended the RNS algorithm by introducing the variable lengths of a detector radius. The authors of this work aim to show an improvement in the detection accuracy and algorithm efficiency, through covering a non-self space with fewer detectors, and cover the holes by using detectors with a smaller radius.

While the RNS algorithm may alleviate the scaling problem of the NS algorithm[3], it creates different problems: the number of detectors required to cover a non-self space and the radius of each detector cannot be estimated in advance, and there is no guarantees of achieving the optimal space coverage with minimum overlap. In order to solve these problems, Gonzalez *et al.* proposed a *randomised* real-valued negative selection (RRNS) algorithm [50][52]. The RRNS algorithm uses *Monte Carlo integration*, which is a well-known randomised algorithm, to calculate the number of detectors needed to cover a non-self space. It first estimates the volume of a self space based on the assumption that the average minimum distances from collected self samples forms the boundary of self space. Then the number of detectors required to cover a non-self space is calculated by defining a fixed length of detector radius, through obtaining the volume of a non-self space as the complementary to the volume of an estimated self space. Furthermore, for the efficiency of detector generation, *simulated annealing* is used to minimise the overlapping spaces covered by detectors. Through this modification the authors show that the RRNS algorithm provides a better non-self space coverage with the same or less computational effort compared to the RNS algorithm. However, scalability tests were not performed with respect to computing time on a realistic size of intrusion related data.

In recent work, Stibor *et al*.[111] gave a comparison between the real-valued positive and negative selection algorithms and two statistical anomaly detection algorithms, the Parzen-Window method and one-class support vector machine(SVM). The comparison revealed that the real-valued positive selection and the Parzen-Window method suffered from long computation times although both produced good classification accuracy. The real-value negative selection algorithm with variable-sized detectors had poor classification performance (the maximum detection rate reached 2.6% on high dimensional space). On the other hand, the one-class SVM produced high detection rates while maintaining an acceptable run-time complexity.

Work in this area is progressing rapidly. While the RNS algorithm attempts to cover a non-self space by generating hyper-sphere detectors, Shaprio *et al*.[102] generated hyper-ellipsoid detetors, which are more flexible in shaping detectors. They used an evolutionary algorithm that reshaped randomly generated hyper-ellipsoid detectors fit to a non-self space. In contrast, the recent work of Zi and Dasgupta[73] attempted to solve the coverage problem by integrating the statistical hypothesis test to the negative selection algorithm. In this approach, the generation of detectors terminates when the hypothesis test rejects the null hypothesis "The coverage of non-self space by all the existing detectors is below an expected percentage". Although this work showed hypothesis test integration with RNS, authors claimed that it can be integrated with any type of negative selection algorithm [73].

---

3 According to Gonzalez [50], further studies are yet required to substantiate this claim fully.

Finally, hybrid approaches that combine NS with other algorithms are becoming more common in recent literature. In [38], Dozier et al. used a steady-state GA to discover the coverage holes of LYSIS type of AIS-IDS. The AIS-IDS (GENERTIA) employed in the following work [68] generates additional detectors that can cover the coverage holes discovered by the steady-state GA. Hang and Dai [60] had a similar motivation but used a different approach. They used anomaly patterns, which are rare but still possible to be collected, as seeds to generate additional synthetic anomalies. Artificial anomalies are generated by using a co-evolutionary GA and the NS algorithm. The co-evolutionary GA abstracts the positive selection process of the HIS, which generalises patterns of the self-class. Then, new artificial anomaly patterns are generated from empty spaces which neighbour a small number of anomaly patterns. The new patterns are then given to the negative selection algorithm with the evolved normal patterns to finalise an artificial anomaly set. The synthetic anomalies can be the coverage holes that might be missed by the LYSIS type of AIS-IDS. These patterns are fed into standard classifiers such as C4.5 and NaiveBayes to detect previously unknown anomalies.

### 3.2.5 Summary and Discussion of Negative Selection Approaches

As seen in this section, various features of the NS algorithm makes it by far the most popular algorithm in solving IDS problems, notably for anomaly detection [46]. However, despite its appealing properties, it has not shown great success in real-life applications. There are two drawbacks to utilising the NS algorithm, namely scalability and coverage, and these are the main barriers to its success as an effective IDS. To tackle the scalability problem, primarily two different types of research efforts have been made. The first group of research is focused on devising more efficient detector generation algorithms. Various approaches have been attempted such as a linear-time algorithm [37], a greedy algorithm that removes redundant detectors [37], [103], [118], and employing diverse ways of evolving detectors [11] [57] [59]. The second group of work has concentrated on employing a new matching function, namely r-chunks matching [14] [42], possibly saving computation time during detector generation and matching. To increase the non-self space coverage of detectors, reducing the number of holes existing in a binary detector space coupled with contiguous matching [62] [64] [75] [13], and proposing real-valued detectors with corresponding matching functions [50] [71] [72] have been investigated. Significant work on a formal framework for positive and negative detection schemes was reported in [42]. This work analyses the trade-offs between two schemes and hence estimates how many self strings are required for either negative or positive detection to ensure that it is computationally advantageous.

However, there are still unresolved issues for the NS algorithm to be an effective IDS. As some researchers argue in [8] [20], possibly the most controversial problem of the NS algorithm is its intrinsic limit starting from the initial assumption - detecting foreign patterns as intrusions. Non-self patterns would not necessarily indicate intrusions and thus a high false positive error rate caused from this assumption is perhaps the inevitable limit of the NS algorithm. Hence, tackling this limit is important future research. The application of a more flexible boundaries between self and non-self space using fuzzy rules [50] [49] is one example of such efforts. However, as pointed out by Stibor [110] [112] et al, there may be inherent problems with the computational efficiency of NS that can never be resolved.

## 3.3 Danger Theory Approaches

Not all AIS are based on the negative selection algorithm (as can be seen by the other major branches in figure 1). Work by Burgess [20] was attentive to alternative immune theories apart from the self and non-self detection model. He claimed that the self and non-self distinction concept, on which the NS based AIS is based, is too simple to explain the whole human immune mechanism, and thus too straightforward to be applied in AIS. Instead, he advocated a different immunology theory called the danger theory [93].

Although this model is still controversial among immunologists, he considered this model to be more appropriate for AIS. The single attribute that makes this model different from other immune theories is that immune response is triggered by unusual deaths of self-cells. Following this model, Burgess [20] [21] put the emphasis of AIS on an *autonomous* and *distributed* feedback and *healing* mechanism, triggered when a small amount of damage could be detected at an initial attacking stage. The system, named *Cfengine*, also aims to automatically configure large numbers of systems on a heterogeneous network with an arbitrary degree of variety in the configuration. After a human administrator initially specifies configuration policies at a very general level using an expert system shell, the system automatically monitors the state of each system and adapts initially specified generic policies to be more locally optimised. This change immediately triggers the modification of other policies affecting different hosts. The new policy reflects a new environment and other hosts can optimise their own policies. Burgess [20] uses an agent framework, employing an expert system based agent that locally optimises the maintenance of each local host in a distributed environment. In [21], Burgess reported that Cfengine has run on an estimated 10,000 nodes around the world since its inception in 1993. He also reported the experiences of Cfengine users as saving administrator's time, scaling well (showing successful running on 2500 Sun hosts) and the minimal load of running Cfengine (recording only a few percent of available CPU time). Burgess explained two main features which contributed to the success of Cfengine: the usefulness of abstract class model and converging traits of its operation. The initial maintenance policy indicates the abstract classes of machines and resources in a network, which is based on several attributes, such as OS, network domain, address and any proposition defined by a user. When Cfengine runs, it applies a configuration policy suitable for the classes of monitoring hosts and resources. The class based generic policy is then locally optimised as Cfengine continues to change the policy depending on what is locally observed. Burgess [24] also formally showed by using game theory that this kind of repeated application of policy change can converge to the stable equilibrium of a local policy. Once the convergent state is reached, Cfengine becomes passive or quiescent until the next considerable anomaly arises. This phenomenon was understood as a form of *system homeostasis*.

The recent development of Cfengine [22] added several new features together with a more sophisticated anomaly detection engine [23] [25] [26]. Within the Cfengine framework, a statistical filter using a time-series prediction detects the significance of deviation. The symbolic content of observed events determines how the system should respond, and is based on a locally optimised policy. Burgess saw a statistical anomaly as a danger signal, and claimed that only the content of observed events characterises the internal degree of the signal. In order to increase the scalability of the anomaly detection component, it incrementally updates the mean and variance of the sampled events. These events are the number of users, the number of processes, average utilisation of the system (load average), number of incoming and outgoing connections based on each service, and the numerical characteristics of incoming and outgoing packets. By performing this mechanism, the model can keep two dimensional time series records, which shows firstly the mean and variance of the current interval (in this case, 5 minutes) and secondly, the mean and variance of a long period (in this case, 2 weeks). Burgess argued that this kind of cross-checking would help to decrease the false positive error rate. Other work by Begnum and Burgess [15] extended Cfengine by combining anomaly detection based automated response mechanism called pH [107]. By the combination of signals from the two systems they intended that pH would be able to adjust its monitoring level based on inputs from Cfengine, and Cfengine would be able to adjust its behaviour in response to signals from pH. Cfengine therefore exploits several features of immune system, most notably being lightweight, self-organised, distributed and multi-layered.

Also influenced by the idea of the danger theory, Le Boudec and Sarafijanovic [101] extended their earlier work on mobile ad-hoc networks [18], [19] [100] (described earlier) and chose to regard a packet loss in the network as a danger signal. In their system the danger signal is used to stop the relevant antigens entering the NS process. When the sequences of protocol events are collected i) at the nodes belonging to the route where the packet loss is observed, and ii) during the time close to the packet loss time, they are considered as non-self antigens. These non-self antigens are not passed to the detector generation process of the NS algorithm. In addition, danger signals are used as co-stimulation signals confirming successful detection through a detector, with good performing detectors becoming memory detectors. Their experiments were carried out on the Glomosin network simulator [121], where 5-20 nodes misbehaved among a total of 40 nodes. The reported test results were firstly, that the use of danger signals strongly impacted on the reduction of false positive error rates and secondly, that the addition of memory detectors also improved detection rates. Once again, their system has the potential to be disposable, distributed, self-organised and light-weight, but has not been demonstrated in a realistic ad-hoc network yet.

Aickelin *et al.* [9] [8] presents the first in-depth discussion on the application of danger theory to intrusion detection and the possibility of combining research from wet and computer laboratory results (see figure 1). They aim to build a computational model of danger theory which they consider important in order to define, explore, and find danger signals. From such models they hope to build novel algorithms and use them to build an IDS with a low false positive error rate. The correlation of danger signals to IDS alerts, and also of IDS alerts to intrusion scenarios, is considered particularly important. Their proposed system collects signals from hosts, the network and elsewhere, and correlates these signals with IDS alerts. Alerts are classified as good or bad in parallel to biological cell death by apoptosis and necrosis. Apoptosis is the process by which cells die as a natural course of events, as opposed to necrosis, where cells die pathologically. It is hoped that alerts can also be correlated to attack scenarios.

To adopt danger signals (apoptosis and necrosis) which trigger artificial immune responses within an AIS, Bentley et al [16] introduced the concept of artificial tissue. The authors stressed that the tissue is an integral part of immune function, with danger signals being released when tissue cells die under stressful conditions. They also highlighted that tissue could play the role of interface between immune responses and pathogenic attacks. The authors argued that the absence of artificial tissue in conventional AIS caused difficulties, with every new AIS needing to be "wired" to a specific problem and hence it was difficult to compare, analyse, and apply such existing AIS to new problems. The authors proposed new tissue growing algorithms designed for AIS that provided generic data representations and hence allowed the artificial tissue to play the role of an interface between a problem and an immune algorithm. The algorithms took a series of input data stream and the artificial tissue grew to form a specific shape by linking input data cells. When new input data was provided to the tissue, the structure of the tissue changed in response. Restructuring tissue caused the deaths of data cells, which released danger signals in return. In this way, the tissue provided a spatial and temporal structure, enabling the AIS to start immune responses which were spatially and temporarily focused. The work exploits immune features such as being self-organised, lightweight and disposable, and has the potential to be implemented as a distributed system, and combined with other immune algorithms to make it multi-layered. To date it has only been tested on UCI machine learning data, however.

Related work by Greensmith et al [58] employed dendritic cells (DC's) within AIS that coordinated T cell immune responses. DCs are a class of antigen presenting cells that ingest antigen or protein fragments in the tissue, and DCs are also receptive to danger signals in the environment that may be associated with antigens. During the antigen ingestion process, immature DC's experience different types of signals that indicate the context (either safe or dangerous) of an environment where the digested antigens exist. The different types of signals lead DCs to differentiate into two different types: mature and semi-mature. The chemical messages known as cytokines produced by mature and semi-mature DCs are different and these different messages influence the differentiation of naïve T cells into several distinctive paths such as helper T cells or killer T cells. In this way, the DC drives the T cell to react to the antigen in an appropriate manner and as such the DC can be seen as the interpretative brain behind the immune responses. Greensmith et al abstracted several properties of DCs that would be useful for application to anomaly detection [58]. In particular, they categorized DC input signals into four groups – PAMPs (signals known to be pathogenic), safe signals (signals known to be normal), danger signals (signals that may indicate changes in behavior) and inflammatory cytokines (signals that amplify the effects of other signals). When each artificial DC experiences the combinations of these four different groups of signals released from the artificial tissue, it interprets the context of ingested antigens by using a signal processing function, which has different weightings depending on the types of input signals. The output of a signal processing function determines the differentiation status of DCs (either semi-mature or mature) and in turn various types of T cells start diverse immune responses. The work is on-going and remains untested for intrusion detection at the time of writing.

Kim et al [87] continued Greensmith et al's work by discussing T cell immunity and tolerance for computer worm detection. Whilst the work of Greensmith et al [58] only presented the role of DCs, Kim et al presented how three T-cell central processes, namely T-cell maturation, differentiation and proliferation would be embedded within the danger theory-based AIS, called CARDINAL (Cooperative Automated worm Response and Detection ImmunNe ALgorithm). CARDINAL attempted to handle three distinctive problems raised in designing the AIS that operates as a cooperative automated worm detection and response system. Firstly, to optimize the number of peer hosts polled, CARDINAL employs the HIS feature that dynamically adjusts the proliferation rate for each effector T cell. Secondly, CARDINAL mimics a T cell differentiation process that leads naïve T cells to mature into various types of effector T cells depending on the interaction with cytokines. This process was expected to provide the AIS with appropriate types of responses depending on the severity and certainty of detected attacks. Finally, CARDINAL was designed to automatically adjust the magnitudes of responses by taking into account both local and peer information. This mechanism was also borrowed from the amplifying and suppressing processes of T cell effector responses via interaction among different types of effector T cells. By proposing CARDINAL, the authors showed how the link between the innate immune system led by DCs and the adaptive immune system operated by T cells. In particular, authors suggested that CARDINAL was not designed to operate in isolation, but in unison as a part of a danger theory inspired AIS, which also employed the artificial tissue [16] and the DC algorithm [58], making a multi-layered system that could exploit most of the other features of the human immune system.

In [86], the artificial tissue, the DC algorithm and T-cell algorithm were combined and presented as a different version of the danger theory inspired AIS. In this work, Kim et al proposed a danger theory inspired AIS for detecting and responding to malicious code execution. The system was designed to provide the ability to i) detect a danger from environmental conditions (by identifying various local anomalies), ii) extract and generalize attack signatures from the data associated with detected danger (by constructing system call policy rules), and iii) respond to an on-going attack appropriately (by permitting or blocking certain system calls presented as the policy rules generated and refined by AIS). The authors anticipated that these abilities would allow the AIS to automatically reconfigure system call policy rules depending on the severity and certainty of attacks. These methods have not been tested for real intrusion detection yet, however.

As shown in [87], the danger theory inspired AIS was designed to adopt both the innate immune system and the adaptive immune system. Apart from the danger theory inspired AIS, most AIS have largely taken their inspiration only from adaptive immune systems while the innate immunity directly links to the adaptive immunity. Twycross and Aickelin [114] provided a review of biological principles and properties of innate immunity, and showed how these could be incorporated into artificial models. In this work, authors addressed six properties of the innate immune system that would influence the capability of AIS. These six properties are summarized in table 1.

**Table 1. General properties of the innate immune system [114]**

| | |
|---|---|
| Property 1 | Pathogens are recognized in different ways by the innate and the adaptive immune systems |
| Property 2 | The innate immune system receptors are determined by evolutionary pressure |
| Property 3 | Response to pathogens is performed by both the innate and adaptive systems |
| Property 4 | The innate immune system initiates and directs the response of the adaptive immune systems |
| Property 5 | The innate immune system maintains populations of adaptive immune system cells |
| Property 6 | Information from the tissue is processed by the innate immune system and passed on to the adaptive immune system |

In addition, the authors evaluated the biological innate immune system based on the conceptual framework proposed by Stepney et al [114]. This framework describes five significant properties that affect complex behaviors in general: openness, diversity, interaction, structure and scale (ODISS). By showing that the innate immune system indeed supports ODISS properties, Twycross and Aickelin highlighted the key and unique role of the innate immune system within an integrated AIS.

## 3.4 Other algorithms

While negative selection and the danger theory are perhaps the most popular approaches in AIS for intrusion detection, there are many researchers who choose to create AIS based on alternative ideas, for practical and philosophical reasons.

In contrast to the negative selection approaches, early work by Somayaji *et al.* [105] aimed to build an IDS based on an explicit notion of self within a computer system (see figure 1). The system was host-based, examining specifically privileged processes. The system collected self-information in the form of root user `sendmail` (a popular UNIX mail transport agent) command sequences. This self-information was constructed as a database of normal commands. Further `sendmail` commands were examined and compared with entries in this database. The authors considered the time complexity for this operation was O(N) where N is the length of the sequence. A command-matching algorithm was implemented and compared with the defined behaviour in the database. Intrusions were detected when the level of mismatched entries in the database had risen above a predefined level. Subsequent alerts were generated, and a basic response was suggested, but no dramatic system changing response was implemented. While this work did exploit the immune properties of being self-organised, it did not make significant use of the other five useful properties of the immune system.

Building on previous work by Somayaji *et al.* [105], the work by Hofmeyr *et al.* [63] was also motivated by the need to improve anomaly-based IDS (see figure 1). Misbehaviour in privileged processes were examined through scrutinising the same superuser protocols, but using a different representation. System call traces were presented in a window of system calls, a value of six selected by a trial-and-error. This window was compared against a database of normal behaviour, stored as a tree structure, compiled during a training period. If a deviation from norm was seen, then a mismatch was generated, with sequence similarity assessed using a Hamming distance metric. A sufficiently high level of mismatches generated an alert, but did nothing to alter the system. All the cases of intrusions tested were detected by the system. With regard to false positives, a bootstrap method was used as a proof of concept, though no actual results were presented. The authors concluded that false positives could be reduced through the increase in the training period. It was claimed that their system was scalable, and generated on average four false positives per day, although they did not directly compare their system with any other. The vast majority of the presented results was evidence of the database scaling well, finding the optimum sequence length and setting the mismatch threshold parameters. The results suggested that this approach could work using data from both real and controlled environments, but the authors found that it was difficult to generate live data in a dynamic environment. They also noted that issues of efficiency had been largely ignored, but would have to be addressed if the system was to work in the real world. Like Somayaji's system, the main immune property exploited was being self-organised.

Stillerman *et al.* [113] built on the work of Hofmeyr *et al.* [63] and introduced an immunity-based intrusion detection approach that was particularly applicable to Common Object Request Broker Architecture (CORBA) applications. CORBA is a popular common messaging middle-ware that enables the communication of distributed objects for distributed applications. The authors employed the same approach reported in [63] to detect a misuse attack performed by a legal user of the system, termed a rogue client attack. The experimental results showed that the system was able to detect anomalies caused by this attack without high false positive error rates. Although this report showed that their work was feasible when applied to the CORBA application level of attack, the report did not include exact false positive error rates or the training and test data collection periods in the experimental results. Furthermore, their work did not utilise any of the new ideas (i.e., negative selection) of AIS, instead directly using a notion of self.

The later work of Ebner *et al.* [39] also employed a notion of self, which identified authorised users by collecting normal typing behaviour of users. The authors chose not to use the NS algorithm, instead simply building normal keystroke profiles using the duration of key presses and the delay between key presses. They claimed that it would be more efficient and easier to detect non-self by building self patterns when a given self space is relatively small, and believed that the identification of self keystroke behaviours belonged to such a case. Melnikov and Tarakanov briefly introduced an immunocomputing model for IDS in [94]. Their model works on a simple test problem, but further results are being produced at the time of writing and may prove promising. Also, Fang and Le-Ping [43] employed the network based AIS, aiNet by De Castro and Von Zuben [27]. aiNet is the clustering algorithm which can be used for anomaly detection by identifying outliers.

Trapnell Jr. [74] proposed a new immune algorithm that removed malicious nodes from a P2P network. His immune algorithm, called the leukocyte-endothelial blacklisting strategy (LEBS), abstracted macrophages, a cytokine called TNF and T cells. In LEBS, machrophage agents move around from one node to another and detect any malicious nodes. Whenever they detect a malicious node[4], they become activated for a limited period and create a number of TNF agents with limited lifetime and the selection expression level. TNF agents are sent out to neighbouring nodes and the selection expression level decays during the lifetime of TNF agents. T cell agents also move around different nodes depending on the selection expression level they experience. They have selection adhesion functions, which take a selection expression level and assign weights to all the neighbor nodes of a node where they locate. The higher weighted the neighbor nodes are, the more likely they are selected as the destination of agents for the next move. T cell agents become activated when they encounter the activated machrophages. Activated T cells then seek for the location of a malicious node presented by the activated machrophages. When they encounter a malicious node as their neighbour nodes, they add this malicious node to the blacklist of a current node, and become unactivated. The LEBS algorithm also mimics innate immunity triggering the proliferation of T cells and provides an effective strategy to distribute the blacklist of malicious nodes to peer nodes within a dynamically changing p2p network. Like the work of Le Boudec and Sarafijanovic [18] [19] [100] reviewed earlier, this application of immune algorithm to a newer type of network architecture may be an excellent approach for these techniques – dynamically changing ubiquitous network environments look set to become more common, and with few existing solutions to intrusion detection, the self-organising, distributed and lightweight AIS may become a popular solution.

# 4 Discussion

In section 2.2, we listed six properties of the immune system that contribute to an effective IDS. The major part of this article has provided detailed overviews of systems proposed and implemented, containing one or more immune-inspired algorithms or concepts. Table 2 summarises the artificial immune algorithms and concepts that the reviewed AISs have employed. It also shows the corresponding biological immune

---

[4] The actual algorithm on how to detect a malicious node is not presented in [74]

features that are expected to be obtained from the implementation of the artificial immune algorithms and concepts.

**Table 2: The Relationship Between Biological Immune Features and Artificial Immune Algorithms**

| Human Immune Features | Artificial Immune Algorithms/Concepts |
|---|---|
| Distributed | Idiotypic Immune Network, Multi-Agent Systems, Negative Selection |
| Multi-layered | Multi-Agent Systems, Co-Stimulation |
| Self-Organised | Gene Library Evolution, Clonal Selection, Negative Selection, Local Sensitivity by Cytokine |
| Lightweight | Memory Cells, Imperfect Detection, Dynamic Cell Turnover |
| Diverse | MHC(Permutation Mask) |
| Disposable | Cell Life Span, |
| Self/Non-Self Detection | Negative Selection, Tolerisation Period |

Some of algorithms mentioned above correspond closely to the classification of immune algorithms by de Castro and Timmis [32] into thymus models, clonal selection algorithms, immune network algorithms and bone marrow models. These correspond to our negative selection, clonal selection, idiotypic network and gene library categories respectively. Table 3 presents the reviewed AISs coupled with the artificial immune algorithms and concepts used. In order to clarify the use of various different types of immune algorithm, we shall concentrate here on 'complete' intrusion detection systems that detect intrusions in real-time, rather than ideas, partial implementations, or simulations.

**Table 3: Summary of immune-based algorithms used by the complete systems**

| AIS | Multi agent | Negative selection | Co-stimulation | Gene libraries | Clonal selection | Local sensitivity | generalised detection | Dynamic cell turnover | Permutation mask | Cell life span | Tolerisation period | Immune memory | Idiotypic immune networks | Response | Self-nonself |
|---|---|---|---|---|---|---|---|---|---|---|---|---|---|---|---|
| Somayaji et al. [105] | | | | | | | | | | | | | | | X |
| Hofmeyr et al. [63] | | | | | | | | | | | | | | | X |
| Hofmeyr et al. [62] | | x | x | | | x | x | x | x | X | x | x | | | x |
| Balthrop et al. [14], [13] | | x | x | | | | x | x | x | x | x | | | | x |
| Kephart et al. [76] [77] [78] | x | | | | | | x | | | | | | | x | |
| Burgess [21] [22] [15] | x | | x | | | | | | | | | | | x | x |

From Table 3 it is evident that the most popular means of implementing an immune system is through the use of a self-nonself model. This approach is used by most systems under review. Only Kephart *et al.* employed a decoy program in detecting new viruses instead of using the self-nonself model. Early work by Burgess [21] [22] did not use the self-nonself model but later work [15] introduced anomaly detectors in order to trigger immune responses, leading to the optimisation of a system administration policy. Four systems out of a total of six were developed by the University of New Mexico team and thus they stem from a similar abstract model of the immune system (see figure 1). The other two systems adopt response and multi-agent mechanisms. No system reviewed here fully implemented an intrusion detection system based on clonal selection, idiotypic networks or gene libraries.

**Table 4: Immune properties provided by the systems listed in table 4.**

| AIS | Distri-buted | Multi-Layered | Self-Organised | Light-weight | Diverse | Disposable |
|---|---|---|---|---|---|---|
| Somayaji et al. [105] | | | x | | | |
| Hofmeyr et al. [63] | | | x | | | |
| Hofmeyr et al. [64] | x | | x | x | x | x |
| Balthrop et al. [14], [13] | x | | x | x | x | x |
| Kephart et al. [76] [77] [78] | | | x | | | |
| Burgess [21] [22] [15] | x | | x | x | | |

Finally, we assess the AISs presented in Table 3 against the immune properties introduced in Table 2. Table 4 illustrates immune features that are provided by complete systems. All of them have self-organising features to an extent, which do not require external management or maintenance in order to automatically detect previously unknown intrusions. None of them provides multi-layered detection. (The multi-layered property was added to other AIS being tested at a small scale [61] [29].) Detector generation through the NS algorithm and the adoption of the permutation mask allows Hofmeyr *et al.*'s [64] and Balthrop *et al.*'s [14], [13] AIS to be disposable and diverse respectively. Several immune mechanisms such as memory cells, imperfect detection and dynamic cell turnover are utilised to create a lightweight detector in Hofmeyr *et al.*'s and Balthrop *et al.*'s system. However, we have also seen that the lightweight property of these systems degrades due to the negative selection process. Burgess's system provides the lightweight feature by adoption of abstract classes of machines and resources.

Examining table 4 further, three systems are distributed (here, we strictly apply the meaning of "distributed" and thus require that there is no central unit maintaining coordination among distributed intrusion processes). However, LYSIS by Hofmeyr *et al.* [64] and Balthrop *et al.* [14], [13] are described as having “intrinsic distributed features”. LYSIS assumes that the system operates under a broadcast LAN environment. A broadcast LAN environment transfers identical input network packets to all the local hosts in a domain. Although LYSIS has different sets of detectors at local hosts, they are exposed to exactly the same set of input network packets. This kind of environment is a very special case. With a switched Ethernet, for example, each host can only experience network packets transferred to it and thus network packets handled by each detector set are different from each other. Due to this rather special circumstance, LYSIS was able to achieve several novel features, such as scalability and robustness, originating from the absence of communication among different detector sets.

From this review it is clear that experimental results so far have shown that *relatively simple* AIS based IDS can work on *relatively simple* problems, i.e. selected test data and small to medium sized testbeds. As shown in this section, there are only six pieces of work that involved the development of IDSs operating in a real-time environment and these works do not fully exploit the many potentially beneficial immune features identified by other researchers. The first obvious direction of future research is that various artificial immune algorithms employing the immune features introduced in table 4 need to be tested and investigated on a much larger scale of real-world environment. In order to show true benefits of immunologically inspired approaches, it is necessary for AIS to be able to detect intrusions on real-time basis.

Another significant future research area is exploring various mechanisms of the human immune system that have not been studied for intrusion detection. Although this survey article clearly shows that some researchers in this field have attempted to adopt new understandings in immunology for intrusion detection, most of work has been carried out based on limited knowledge on the immune system. Considering that new discoveries and understanding in the human immune system are constantly announced, it is important to embed better understood immune models to artificial systems for their success. As shown in [8],[9] which is research conducted in collaboration with immunologists, the latest discoveries in immunology can provide an informative insight on designing a completely new model of AIS. Perhaps a revolutionary solution to computer security problems is more likely if we can employ a revolutionary new understanding in the human immune system.

In summary, this review is the first to be carried out on the rapidly-growing area of AIS for IDS. Our analysis of the literature shows that progress is being made in a wide diversity of areas, but that there is a clear need for research to focus on:

- the incorporation of up-to-date immunobiological findings in immune models to ensure that immune algorithms produce real advantages;
- the exploitation of all known useful features of the immune system in such systems (e.g., being distributed, self-organised, lightweight, multi-layered, diverse and disposable);
- the production of systems that will scale to real-world network environments;
- and the investigation of immune solutions in novel network architectures where no conventional solutions for intrusion detection currently exist (for example an artificial immune solution to intrusion detection seems like an ideal choice in the rapidly growing area of ubiquitous computing, which employs novel computing environments such as MANET, sensor networks, P2P, and mesh networks).

# 5 Conclusion

The analogy between the HIS and IDS naturally attracts computer scientists to research on immune system approaches to intrusion detection. An increasing amount of work has been published on this topic recently and here we have collated the algorithms used, the development of the systems and the outcome of their implementations. The review conducted in this paper focused on providing an overview of IDS for AIS researchers to identify suitable intrusion detection research problems. Information was also provided for IDS researchers about current AIS solutions. We have summarised six immune features that are desirable in an effective IDS: distributed, multi-layered, self-organised, lightweight, diverse and disposable. In addition, we have provided a comprehensive phylogeny of artificial immune algorithms and concepts that have been proposed and implemented in previous work. These artificial immune algorithms and concepts were shown to provide six desirable immune features.

Through careful examination of literature presented in this paper, one can conclude that immunologically-inspired IDS still have much room to grow and many areas to explore. The phylogenetic tree given in figure 1 clearly illustrates that the history of research in this area has shown a clear focus on three major ideas:

1. methods inspired by the immune system that employ conventional algorithms, for example, IBM's virus detector [76]
2. the negative selection paradigm as introduced by Forrest [106][45]
3. approaches that exploit the Danger Theory [93]

with younger methods based on alternative approaches still being developed.

Will larger scale implementations borrowing more heavily from the HIS, i.e. by incorporating aspects such as idiotypic networks, gene libraries and danger theory, be successful? Will immune algorithms implemented in ubiquitous computing environments become mainstream solutions in the future? Such work is currently underway by [8] and others. The proof is yet to come, but if it works *in vivo*, we ought to be able to make it work *in silico*!

# Acknowledgements

This project is supported by the EPSRC (GR/S47809/01), Hewlett-Packard Labs, Bristol, and the Firestorm intrusion detection system team.

**Appendix 1 Glossary and abbreviations of commonly used terms.**

| Term | Abbreviation | Meaning |
|---|---|---|
| Intrusion detection system | IDS | Software systems which identify misuse of computer networks. |
| Misuse detection | - | Intrusion pattern matching algorithm |
| Anomaly detection | - | Detection of difference from a learned norm |
| False positive | FP | An error, where normal is wrongly identified as an intrusion |
| False negative | FN | An error, where an intrusion is wrongly identified as normal |
| Human Immune System | HIS | The cells and processes that protect us against harmful pathogens. |
| Artificial Immune Systems | AIS | The range of algorithms (e.g. negative selection) based on inspiration from the human immune system |
| Antigen | Ag | In biology, the molecule used by immune cells to identify pathogens; in AIS, the datum analysed during the detection process. Self-antigens comprise normal data; non-self antigens comprise abnormal data (potentially representing a pathogen or intrusion). |
| Antibody | various (Ig, IgG, IgA, IgM, IgD, IgE) | In biology, a protein produced by B-cells that is designed to stick to specific antigens, in AIS, often used interchangeably with "detector" and sometimes confused with T-cells or B-cells. |
| Danger Theory | DT | Theory of HIS that suggests harmful pathogens can be detected by examining "danger signals" generated by cells killed abnormally by pathogens. |
| Negative Selection | NS | Theory of HIS that suggests the immune system performs |
| | | anomaly detection by creating detectors that match everything except “self antigens”. |
| Dendritic cell | DC | An important class of antigen presenting cells in the HIS that ingest antigens or protein fragments and that are receptive to danger signals. |
| B-cell | - | Important immune cells in the HIS that produce antibodies. |
| T-cell | various (naïve, Th, Th1, Th2, CTL) | Significant immune cells in the HIS that differentiate into different classes when mature, such as helper T (Th) and killer T cells (CTL). T cells primarily detect intracellular pathogens that escape antibody detection. |